\documentclass{article}
\pdfoutput=1
\usepackage{microtype}
\usepackage{graphicx}
\usepackage{subfigure}
\usepackage{booktabs} %

\usepackage{hyperref}

\usepackage[accepted]{icml2024}

\usepackage{amsmath}
\usepackage{amssymb}
\usepackage{mathtools}
\usepackage{amsthm}
\usepackage{placeins}

\usepackage[capitalize,noabbrev]{cleveref}

\theoremstyle{plain}

\theoremstyle{definition}

\theoremstyle{remark}

\usepackage[textsize=tiny]{todonotes}

\usepackage{tikz}
\usepackage{varwidth}
\usetikzlibrary{decorations.pathreplacing}
\usetikzlibrary{fit}
\usetikzlibrary{positioning}
\usetikzlibrary{calc} 

\usepackage[binary-units=true]{siunitx}

\usepackage{csquotes}

\usepackage{tabularx}
\usepackage{makecell}

\usepackage{graphicx}
\usepackage{subfigure}
\usepackage{algorithm}
\usepackage{algorithmic}

\usepackage{xspace}

\usepackage[commandnameprefix=always]{changes}

\newcommand{\systemName}{MicroNAS\xspace}

\newcommand{\nucleol}{Nucleo-L552ZE-Q\xspace}
\newcommand{\nucleof}{Nucleo-F446RE\xspace}

\icmltitlerunning{\systemName}

\begin{document}

\twocolumn[
    \icmltitle{\systemName: Memory and Latency Constrained Hardware-Aware Neural Architecture Search for Time Series Classification on Microcontrollers}

        \begin{icmlauthorlist}
        \icmlauthor{Tobias King}{yyy}
        \icmlauthor{Yexu Zhou}{yyy}
        \icmlauthor{Tobias Roddiger}{yyy}
        \icmlauthor{Michael Beigl}{yyy}
    \end{icmlauthorlist}
    \icmlaffiliation{yyy}{Karlsruhe Institute of Technology, Karlsruhe, Germany}

    \icmlcorrespondingauthor{Tobias King}{tobias.king@kit.edu}
    \icmlcorrespondingauthor{Yexu Zhou}{yexu.zhou@kit.edu}
    \icmlcorrespondingauthor{Tobias Roddiger}{tobias.roeddiger@kit.edu}
    \icmlcorrespondingauthor{Michael Beigl}{michael.beigl@kit.edu}

    \icmlkeywords{Machine Learning, ICML}

    \vskip 0.3in
]

\printAffiliationsAndNotice{}

\begin{abstract}
    Designing domain specific neural networks is a time-consuming, error-prone, and expensive task.
    Neural Architecture Search (NAS) exists to simplify domain-specific model development but there is a gap in the literature for time series classification on microcontrollers.
    Therefore, we adapt the concept of differentiable neural architecture search (DNAS) to solve the time-series classification problem on resource-constrained microcontrollers (MCUs).
    We introduce \systemName, a domain-specific HW-NAS system integration of DNAS, Latency Lookup Tables, dynamic convolutions and a novel search space specifically designed for time-series classification on MCUs.
    The resulting system is hardware-aware and can generate neural network architectures that satisfy user-defined limits on the execution latency and peak memory consumption.
    Our extensive studies on different MCUs and standard benchmark datasets demonstrate that \systemName finds MCU-tailored architectures that achieve performance (F1-score) near to state-of-the-art desktop models.
    We also show that our approach is superior in adhering to memory and latency constraints compared to domain-independent NAS baselines such as DARTS.

\end{abstract}

\section{Introduction}
MCUs are small, low-power computing systems that can be found in a wide range of devices, including medical equipment, consumer electronics, wearables and many more.
Deploying machine learning models directly on microcontrollers enables applications such as predictive maintenance \cite{predictiveMaintenance}, human activity recognition \cite{humanActivityRecognition} or health monitoring \cite{healthMonitoring} to be always available without network connectivity while ensuring privacy \cite{edgePrivacy}.
Many of these devices utilize sensors, such as, accelerometers, gyroscopes and more which generate time series data \cite{smartSensors}.

The combination of sensors and microprocessors embedded in smart sensors creates the opportunity for offline, on-device data analysis which allows these devices to operate in privacy-critical, real-time and autonomous systems \cite{smart_sensors}.
Due to the limited hardware of typical MCUs (e.g. \SI{64}{\kilo\byte} SRAM, \SI{64}{\mega\hertz} CPU clock),
it is not possible to run state-of-the-art time series classification architectures such as InceptionTime \cite{InceptionTime} or DeepConvLSTM \cite{DeepConvLSTM} on these devices.

A common solution to deal with the limited resources of microcontrollers is to send the raw data to a server in the cloud, where state-of-the-art models can be executed and then transmit the result back to the microcontroller.
For many reasons, this approach is not sustainable: network communication introduces uncertain latencies to the system preventing its use in real-time applications or scenarios where networking is not available, processing data on external servers creates a privacy risk. In addition, network communication is expensive for microcontrollers in terms of energy consumption.
Another option is to manually design specific neural networks for individual use-cases.
This is often done by domain experts with knowledge in the field of machine learning and is an error-prone and time-consuming process \cite{towards_autoTune_networks_mendoza}.
To automate this design process, neural architecture search (NAS) can be applied to find suitable neural network architectures for specific use-cases.
Existing state-of-the-art NAS-systems focus on generating neural network architectures for image classification \cite{darts, FBNet, fbnetv2}.
Hardware-aware NAS systems (HW-NAS) which optimize classification accuracy and hardware utilization have also been implemented for image classification  \cite{micronets, uNAS_liberis, ProxylessNAS}.
However, existing HW-NAS systems are not adapted to the time series classification task and utilize latency estimation methods that are not precise enough for highly constrained microcontrollers \cite{micronets, uNAS_liberis, fbnetv2}.

To apply HW-NAS to time series classification, two main challenges need to be overcome.
First, the shape of time series data differs fundamentally from image data which requires an adaptation of the search space.
We solve this problem by introducing a novel, two stage search space, in which first Time-Reduce cells extract temporal context and in a second step, Sensor-Fusion cells allow for cross-channel interaction \cite{tinyHar}.
Depending on the window-size and the number of sensor-channels, we vary the number of cells in the search space to cover a wide range of time series datasets.
Second, to be able to adhere to the resource constraints of MCUs and select the best architecture, a fine granular search space in combination with precise execution latency predictions is required.
If the search space is to coarse, it may not be possible to find optimal architectures for the given task that still satisfy user imposed limits on the execution latency and peak memory consumption.
Similarly, imprecise execution latency estimations make it impossible to determine when the maximum allowed execution latency is exceeded.
We utilize a masking convolution approach adapted from \cite{fbnetv2} to create a fine granular search space by varying the number of filters in convolutional layers.
To precisely estimate the execution latency of architectures in the search space, we employ a latency lookup table based approach \cite{FBNet}.
\citet{fbnetv2} employ a technique called effective-shape-propagation in order to estimate the execution latency of architectures.
This approach is not compatible with the lookup-table based approach but we overcome this limitation by linking these two techniques with an interpolation schema.
In summary, this paper makes the following contributions:
\begin{enumerate}
    \item \systemName; the first hardware-aware neural architecture search (HW-NAS) system for time series classification tasks on embedded microcontrollers.
    \item Introduction of a time series classification specific search space suitable for datasets with varying window sizes and number of sensors. The search space contains two searchable cells that extract temporal information and allow for cross-channel interaction respectively.
    \item An automatic characterization method to calculate neural architecture execution latencies for microcontrollers based on a lookup table with an average error of $\approx \pm \SI{1.59}{\milli\second}$, showing that this approach outperforms proxy latency metrics ($\approx \pm \SI{15.57}{\milli\second}$).
\end{enumerate}

\begin{figure*}[tbp]
    \centering
    \includegraphics[page=6, width=0.8\linewidth]{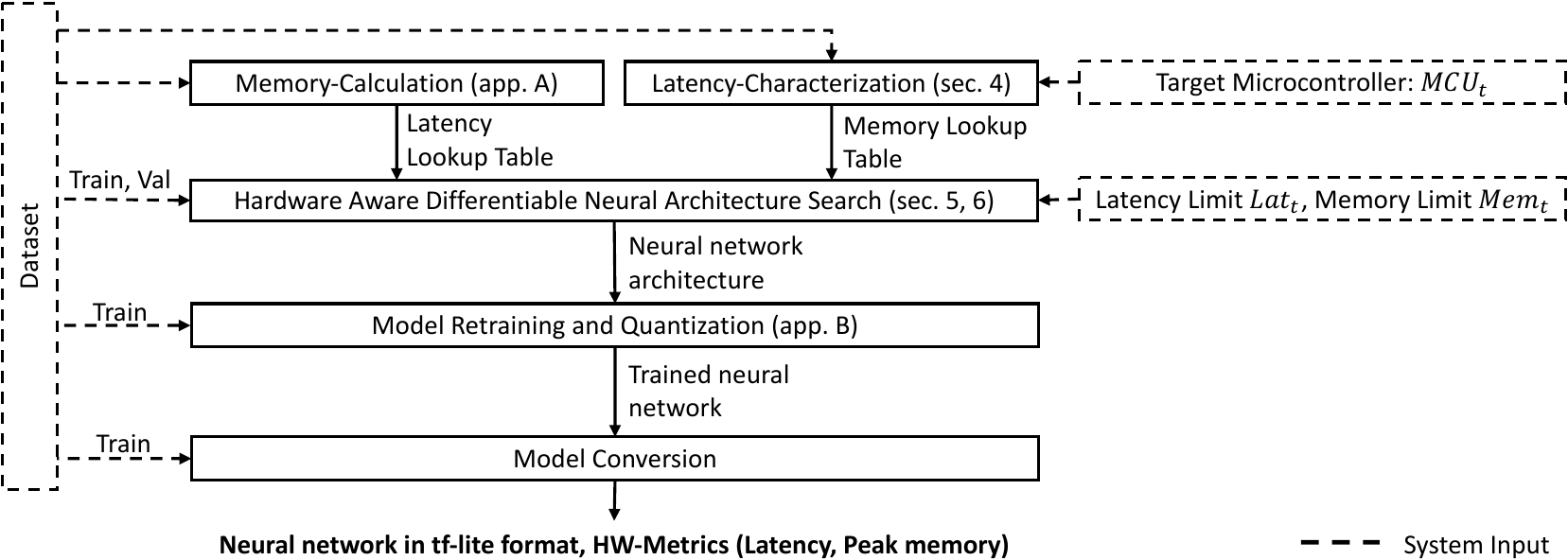}
    \caption{
    \systemName requires the dataset to be split into three different sets which are used at different stages in the pipeline. The user specifies the dataset to be used, the target MCU ($MCU_t$) and the maximum allowed hardware utilization in terms of execution latency ($Lat_t$) and peak memory consumption ($Mem_t$). Output of the system is a corresponding neural network in the tf-lite format.
    }
    \label{fig:system_overview}
\end{figure*}

\section{Background and Related Work}
We first summarize time-series classification using deep learning approaches and then introduce existing state-of-the-art neural architecture search systems.

\subsection{Time Series Classification}
\label{subsec:tsc}
In the recent past, deep learning based approaches have been used successfully for time series classification.
Systems such as InceptionTime \cite{InceptionTime} and the system by \citet{multi_scale_cnn} feature CNN based architectures to aggregate temporal context on multiple scales.
Other options include the use of RNNs or hybrid models consisting of both CNN and RNN layers \cite{skoda_mahmud, DeepConvLSTM}.
Due to computational complexity, using such systems on MCUs is not feasible. Therefore, neural networks specifically designed for the time-series classification task on microcontrollers must be developed \cite{ts_mcu_001, ts_mcu_002}.
Inspired by the existing CNN system architectures, we develop our NAS search space.
This space features searchable cells that are tailored for MCUs and, at the same time, can represent typical structures found in CNN time series classification systems.

\subsection{Neural Architecture Search}
\label{subsec:nas}
Early neural architecture search systems (NAS) \cite{rl_nas_zoph} formulate the search as a reinforcement learning problem.
While this approach produces novel, well-performing architectures, the search takes long as each iteration of the REINFORCE-algorithm requires training a neural network until convergence with no weight sharing between architectures.
To overcome this issue, super-networks have been introduced as search spaces, where each architecture exists as a subgraph, allowing for shared weights among them \cite{darts, enas}.
\citet{smash} and \citet{enas} show, that training the super-network is enough to emulate any architecture in the search space.
\citet{darts} extend this idea by introducing Differentiable Neural Architecture Search (DNAS).
DNAS utilizes a relaxation schema to make the search continuous, differentiable and, therefore, more resource efficient.
In DNAS, the search space is also defined by a super-network, in which a layer has not one but multiple operations.
The layer output $l$ is then computed as a convex combination of the output of the operations $o$ scaled by the architectural weights $\alpha$: $l = \sum_{i} o_i * \alpha_i$.
During architecture search, the regular neural network weights and the architectural weights are jointly optimized using gradient descent.
This allows for a structured and more efficient search.
After training is complete, the architectural weights are used to identify the selected architecture.
Due to its efficiency, we use DNAS as the basis for the search algorithm of \systemName.

Recently, NAS has been extended to be hardware aware (HW-NAS).
Systems in this category not only optimize for classical performance metrics such as accuracy or precision but also for hardware specific metrics such as execution latency, peak memory and energy consumption \cite{survey_benmeziane}.
Optimizing the hardware utilization is especially important when targeting microcontrollers as these devices are typically severely resource constraint.
Therefore, during architecture search time, the search algorithm needs to be able to estimate relevant hardware metrics for arbitrary architectures.
For the peak memory consumption, analytical estimation can be used for precise calculation.
In contrast, for the execution latency, many approaches exist \cite{survey_benmeziane}.
Real-time latency measurements on the target hardware during architecture search provide precise measurements but prolong the search drastically \cite{survey_benmeziane}.
Another common and much faster approach is to use the number of flops or similar metrics as a proxy for the execution latency \cite{FBNet, uNAS_liberis, micronets}.
While the authors of MicroNets \cite{micronets} and $\mu$Nas \cite{uNAS_liberis} claim the number of operations in a model to be a good proxy for the execution latency when targeting MCUs, \citet{not_all_ops_are_created_equal} argue that this is not the case. 
A middle ground between the slow but precise on-device measurements during search and the fast but imprecise latency estimations using the number of operations, are lookup tables \cite{survey_benmeziane}.
With the lookup table approach, operations in the search space are executed on the MCU once and can then be efficiently used during search time.

Existing NAS approaches that target time series data are concerned with classification \cite{tsNas_rakhshani} and forecasting \cite{tsNAS_chen} but do not target MCUs and are not hardware aware.
In contrast, HW-NAS systems which target MCUs are not concerned with time-series classification \cite{micronets, uNAS_liberis}. 
This underlines the need for a HW-NAS time series classification system that combines the techniques of differentiable neural architecture search and the lookup table based latency estimation approach.

\section{System Overview}
The system overview of \systemName can be seen in \autoref{fig:system_overview}. 
The input to the system consists of a time series dataset, an MCU to use ($MCU_t$) and user-defined limits on the execution latency ($Lat_t$) and peak memory consumption ($Mem_t$).
In a first step, the hardware utilization of each operator in the search space is obtained in an operation called characterization, shown in \autoref{sec:characterization}.
After characterization, HW-NAS is executed where a DNAS approach (\autoref{sec:system_search_algorithm}) is utilized to select a suitable architecture from our search space (\autoref{sec:system_search_space}) for the dataset and $MUC_t$ combination.
The found architecture is then extracted from the search space and retrained from scratch using quantization aware training to maximize classification performance (\autoref{sec:model_retraining_quantization}).
Finally, the trained, $int\_8$ quantized neural network is converted to the tf-lite format and can now be deployed on $MCU_t$.

\section{Latency \& Peak Memory Estimation}
\label{sec:characterization}
For \systemName to find architectures which obey user-defined limits on the execution latency and peak-memory consumption, it is necessary to estimate the actual execution latency and peak-memory consumption of individual architectures in the search space.
To improve on flops-based proxy metrics for the execution latency, we introduce a lookup-table based approach.
In \autoref{sec:peak_mem} we then outline how to analytically estimate the peak-memory consumption as previously done by \cite{micronets, uNAS_liberis}.

\subsection{Latency Characterization}
\label{sec:execution_latency}
\begin{figure}
        \centering
        \includegraphics[width=\linewidth]{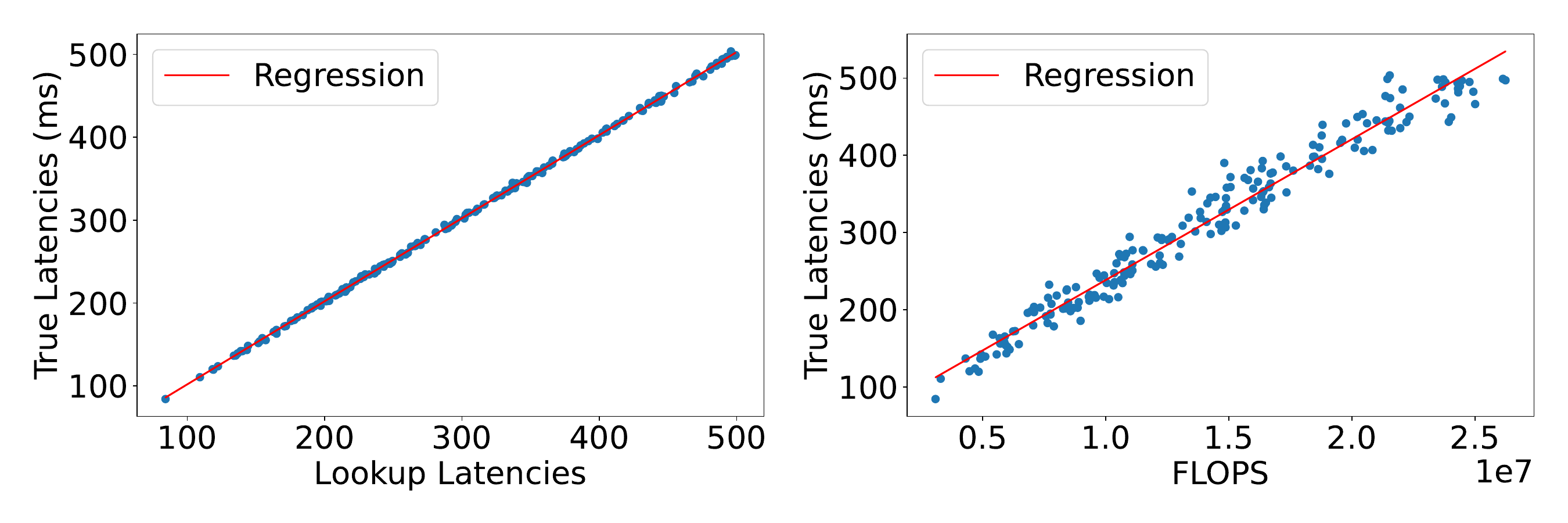}

        \caption{Execution latency of whole architectures from our search space.
        Left: Our lookup-table latency approach. MAE: \SI{1.59}{\milli\second}, $R^2$: \SI{99.97}{\percent}.
        Right: Flops based estimate: MAE: \SI{15.57}{\milli\second}, $R^2$: \SI{96.78}{\percent}.}
        \label{fig:ops_vs_latency}
\end{figure}

When calculating the execution latency of neural network architectures, the literature proposes to use the number of operations in an architecture as a proxy metric \cite{uNAS_liberis,micronets,liberies_pruning}. 
We argue for a lookup table based approach, in which we obtain the execution latency of each operator in our search space by executing it on the actual MCU.
From this information, we can calculate the execution latency for arbitrary architectures in the search space.
To determine the viability of this approach, we conduct our own experiment in which we compare our latency lookup table approach with a flops-based proxy metric as seen in \autoref{fig:ops_vs_latency}.
We executed the experiment using $int\_8$ quantized networks on the Nucleo-L552ZE-Q where we measure the actual execution latency by using the internal CPU-cycle counter on ARM-Cortex processors.
Results can be seen in figure \ref{fig:ops_vs_latency}.
Our lookup table approach achieves an $R^2$-score of $99.97$ with a mean absolute error of $1.59$ ms.
The flops based latency estimation achieves an $R^2$-score of $96.78$ and a mean absolute error of $15.57$ ms.
Therefore, we can conclude, that the lookup table approach is able to outperform the flops-based latency estimation approach.

\section{Search Space}
\label{sec:system_search_space}
\begin{figure*}
    \centering
    \includegraphics[page=8, width=0.8\linewidth]{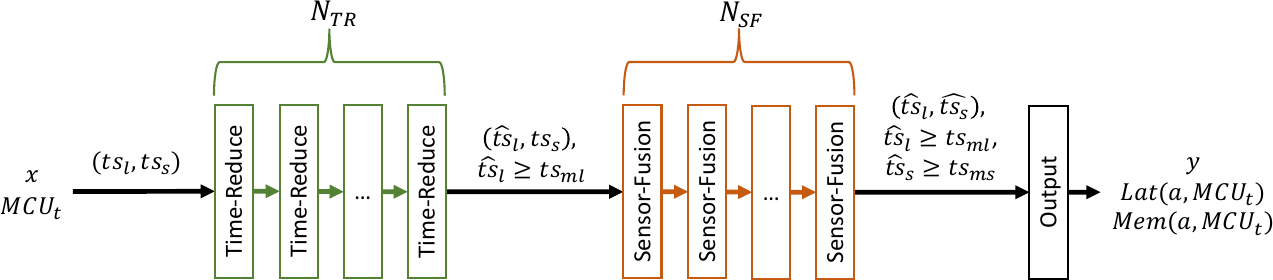}
    \caption{High-level overview over the search space. The raw, windowed time series $x$ with shape $(ts_l, ts_s)$ is propagated though $N_{TR}$ Time-Reduce and $N_{SF}$ many Sensor-fusion cells. The resulting time series is then of shape $(\hat{ts}_l, \hat{ts}_s)$. Class probabilities $y$ and hardware metrics are output by the Output cell at the end of the network.}
    \label{fig:macro_arch_overview}
\end{figure*}

To accommodate the time-series classification task, we introduce a novel, MCU-tailored search space consisting of two types of architecture-searchable cells.
This search-space is defined by a super-network, build from a linear stack of architecture-searchable cells.
To support the time series classification task, two types of searchable cells are designed.
First, Time-Reduce cells are utilized to extract temporal context from the incoming time series.
In a second step, Sensor-Fusion cells allow for cross-channel interactions where information from multiple sensors can be fused.
This two-step process is a common approach in the domain of time series classification \cite{multi_scale_cnn, DeepConvLSTM, TSCMultiConv, attendAndDiscriminate,  tinyHar}.
Each of the searchable cells is hardware aware and therefore output their hardware metrics, the execution latency $Lat(\alpha, MCU_t)$ and the peak memory consumption $Mem(\alpha, MCU_t)$ which depend on the architectural weights $\alpha$ and $MUC_t$.
To adapt the search space dynamically to datasets with varying window sizes ($ts_l$) and number of sensor-channels ($ts_s$), the number of cells is adapted automatically.
The number of Time-Reduce cells is calculated according to:
$$N_{TR} = \left\lfloor log_2(\frac{ts_{l}}{ts_{ml}}) \right\rfloor$$
and the number of Sensor-Fusion cells is calculated according to:
$$ 
N_{SF} = \left\lfloor \log_2\left(\frac{ts_{s}}{ts_{ms}}\right) \right\rfloor \left(1 + sf_{s}\right)
$$
$ts_{ml}$ is the minimum window size after the Time-Reduce cells while $ts_{ms}$ is the minimum size of the sensor-dimension after the Sensor-Fusion cells.
The parameter $sf_s$ is user settable to increase the number of Sensor-Fusion cells which allows for deeper networks.
The Sensor-Fusion cells can be configured with stride $1$ or $2$ while the number of cells with stride $2$ is independent of the parameter $sf_s$.
An overview of this search-space can be seen in \autoref{fig:macro_arch_overview}.
To improve stability during training, dropout layers with dropout factor $0.3$ are placed between all cells (not shown in figure).

\subsection{Decision Groups}
\label{sec:decision_group}
In DNAS, each searchable cell contains two sets of weights: The regular neural network weights $w$ as well as the architectural weights $\alpha$ indicating the architecture.
These architectural weights are organized in decision groups.
A decision group $\alpha_i$ is a collection of architectural weights $\alpha_{i,j}$, used to make one-out-of-many decisions.
$\alpha_{i,j}$ denotes the $j$-th architectural weight in the $i$-th decision group.
Each weight $\alpha_{i,j}$ in a decision group gates a path in the cell and therefore, one-hot encoded decision groups define the cell-architecture.
During search-time, a pseudo probability-function is applied to the decision group:
\begin{equation}
    \Hat{\alpha}_{i,j} = pseudo\_prob(\alpha_{i,j})
\end{equation}
After the search, each decision group will be one-hot encoded.
This effectively discards all options which are assigned the zero value and therefore the final architecture is determined.

\subsection{Dynamic Convolutions}
\label{sec:dyn_conv}

A dynamic convolution \cite{fbnetv2} is a convolution whose number of filters can be searched for efficiently by using weight sharing.
We adapt this concept and couple it with an interpolation schema to  make it compatible with our latency lookup table.
In a dynamic convolution, first a convolution with the maximum number of allowed filters $f_{max}$  is applied to the input $x$.
The output of this convolution is then multiplied with a binary mask in the filter-dimension.
This mask is the weighted sum of several masks $m_i$, with architectural weights $\hat{\alpha}_{y,i}$:
$$y = conv(x) * (\sum_{i}^{}{\hat{\alpha}_{y, i} * m_i})$$
This formulation allows to efficiently search for the number of filters in a convolution by using the decision group $\alpha_{y}$.
As the hardware utilization of a convolution also depends on the number of filters in the incoming time series, we need to take the decision group $\alpha_{x}$, responsible for the number of filters in the input into consideration.
To reduce the cost of latency characterization, we introduce the granularity $g$ with ($f_{max} \mod g == 0$).
This parameter controls how many filters are disabled by one mask $m_i$.
To characterize a dynamic convolution, we must execute all possible combinations of number of input and number of output filters on the $MCU_t$.
The introduction of $g$ reduces the number of possible combinations from $f_{max}^2$ to $(f_{max}/g)^2$ which significantly reduces characterization cost.
Finally, execution latency and peak memory consumption for a dynamic convolution can be calculated with the interpolation schema according to \autoref{eq:dyn_op_lat}.
In the equation, the function $HW(x,y)$ returns the execution latency and peak memory consumption for the dynamic convolution with $x$ input and $y$ output filters.

\begin{equation}
    \begin{gathered}
        op_{hw} = \hat{\alpha}_{y}^T \cdot HW_{op} \cdot \hat{\alpha}_{x} \\
        \text{with} \\
        HW_{op} = \resizebox{0.7\linewidth}{!}{
            $\begin{bmatrix}
                HW(g_x, g_y)                      & \dots  & HW(\vert\alpha_{x}\vert g_x, g_y)                      \\
                \vdots                            & \ddots & \vdots                                                 \\
                HW(g_1, \vert\alpha_{y}\vert*g_y) & \dots  & HW(\vert\alpha_{x}\vert g_x, \vert\alpha_{y}\vert g_y)
            \end{bmatrix}$
        }
    \end{gathered}
    \label{eq:dyn_op_lat}
\end{equation}

\begin{figure}
    \centering
    \includegraphics[width=0.7\linewidth, page=9]{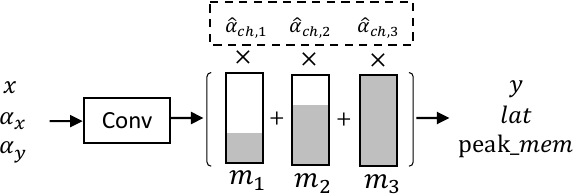}
    \caption{Dynamic convolution with three different options (e.g. $f_{max} = 24$, $g = 3$) for the number of filters. The binary masks ($m_i$) zero out certain filters in the output of the convolution. Grey areas are ones and white areas are zeros.h}
    \label{fig:dyn_conv}
\end{figure}
In the equation, $g_y$ denotes the granularity corresponding to the output of the convolution while $g_x$ corresponds to the granularity of the input.
The same concept can be applied to the dynamic Add operations.

\subsection{Cells}
\label{sec:search_cells}
To accommodate the time series classification task, two types of searchable cells are designed. The Time-Reduce cell aggregates information in the temporal domain while the Sensor-Fusion cell allows for cross channel interaction. After each convolution in the architecture, a ReLU activation function is applied (not shown in graphics).

\subsubsection{Time-Reduce Cell}
\begin{figure}
    \centering
    \includegraphics[width=1\linewidth, page=1]{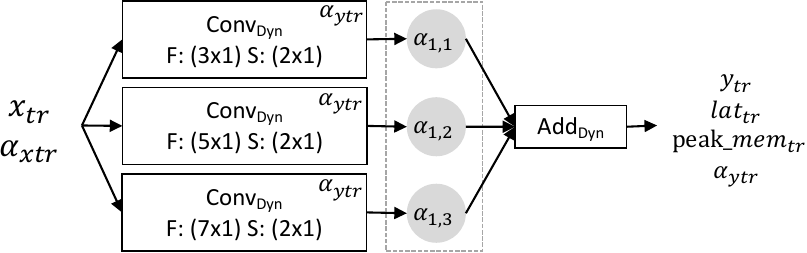}
    \caption{Time-Reduce cell. Contains two decision groups. $\alpha_1$ to choose a convolution and $\alpha_{ytr}$ to search for the number of filters. $F$ is the filter size while $S$ is the stride configuration.}
    \label{fig:time-reduce-cell}
\end{figure}

The Time-Reduce cell shown in \autoref{fig:time-reduce-cell} aggregates local context by applying strided convolutions in the temporal dimension while leaving the sensor-dimension untouched (Filter size: $(\{3,5,7\} \times \bf 1)$). This is done to reduce the window size of the propagated time series, to save on computational costs in subsequent cells but also to extract and fuse local initial features from the raw data \cite{tinyHar}.
The cell contains two decision groups.
$\alpha_1$ to choose one of the convolutions and $\alpha_2$ to select the number of filters in that convolution.
Input to this type of cell is a time series $x_{tr}$ of shape $(t_{in}, s_{in}, f_{in})$ while the output $y_{tr}$ is of shape $(t_{out}, s_{out}, f_{out}) = (0.5 \times t_{in}, s_{in}, f_{in})$
The cell also receives the decision group $\alpha_{xtr}$ indicating the number of filters in $x_{tr}$.

\subsubsection{Sensor-Fusion Cell}
\begin{figure}
    \centering
    \includegraphics[width=\linewidth, page=11]{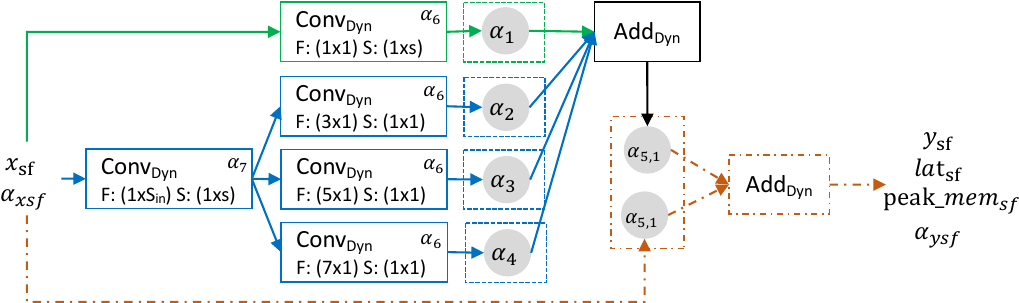}
    \caption{Sensor-Fusion cell. Consists of three pathways, can be configured with stride one or two and depending on that contains six or seven decision groups. $F$ denotes the filter size while $S$ denotes the stride configuration. The orange pathway is only active, when $stride_{sf} = 1$.}
    \label{fig:Sensor-Fusion-cell}
\end{figure}

A common problem when dealing with time series data is the interaction between the different sensors \cite{tsc_bakeoff, tinyHar}.
To tackle this problem, the Sensor-Fusion cell, inspired by InceptionTime \cite{InceptionTime}, seen in \autoref{fig:Sensor-Fusion-cell} was designed.
Input to the cell is a time series $x_{sf}$ of shape $(t_{in}, s_{in}, f_{in})$.
The cell can be configured with stride $stride_{sf}$ to be equal to $1$ or $2$ which influences the output shape to be $(t_{in}, s_{in} / stride_{sf}, f_{in})$.
When the stride equals one, three pathways through the layer exist, shown in green, blue and orange.
The orange pathway (dashed) is an identity connection which can be used to skip the layer and is only included if the input and output of the layer have the same shape and is therefore omitted when the stride equals $2$.
In the main pathway through the cell (shown in blue), first, a dynamic convolution with filter size $(1, S_{in})$ is applied to allow cross channel interaction by performing convolution across all sensor-channels.
Then, in a next step, multiple convolutions with filter sizes $(f,1), \, f \in \{3,5,7\}$ are applied. Each of these convolutions can be individually turned on or off by the search algorithm using the decision groups $\alpha_{2,3,4}$.
This allows features to be extracted simultaneously at different temporal scales if necessary.
In figure \ref{fig:Sensor-Fusion-cell}, these decision groups are drawn with only one weight, although in reality, for each of the three convolutions, a second parallel zero-connection exists as an alternative which allows the individual selection process.
In addition, a skip-connection (shown in green) can be added to the layer using $\alpha_1$.
As all the pathways through the cell must output tensors with the same shape, the dynamic convolutions in the skip-connection and the dynamic convolutions in the main-block share their decision group $\alpha_6$ to select the number of filters.

\subsubsection{Output cell}
\begin{figure}
    \centering
    \includegraphics[width=\linewidth, page=4]{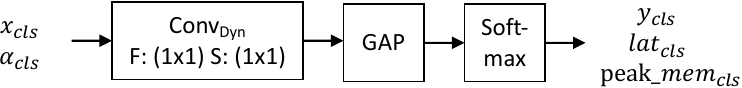}
    \caption{The output cell features a fixed architecture and is therefore not searchable. First, a dynamic convolution is applied to deal with the different number of channels in $x_{cls}$. Finally, global average pooling and a Softmax operation are utilized to output the class probabilities.}
    \label{fig:cls-layer}
\end{figure}

The output cell as seen in \autoref{fig:cls-layer} has a no learnable architecture.
It consists of a dynamic convolution where the number of filters is fixed to the number of classes.
Finally, class probabilities ($y_{cls}$) are output using a Global Average Pooling (GAP) and Softmax layer.

\section{Search Algorithm}
\label{sec:system_search_algorithm}
To search for a suitable architecture in the search space, we apply a modified version of the DNAS algorithm introduced in DARTS \cite{darts}.
We adapt the algorithm to be hardware aware using a multi-objective loss to optimize the architectural weights $\alpha$.
To force the individual decision groups to converge, we employ the Gumbel-Softmax-function \cite{gumbel_softmax} with decreasing temperature $\tau$ as the $pseudo\_prob$-function.
Therefore, we optimize the architectural weights $\alpha$ using the loss-function shown in \autoref{eq:loss_function}.
\begin{equation}   
    \begin{aligned}
        &\mathcal{L}(\alpha, w, MCU_t, Lat_t, Mem_t) = \\
        &loss_{val}(\alpha, w) + loss_{lat}(\alpha, MCU_t, Lat_t) \\
        &+ loss_{mem}(\alpha, MCU_t, Mem_t)
    \end{aligned}
    \label{eq:loss_function}
\end{equation}
In the equation, $loss_{val}$ denotes the cross-entropy loss on the validation dataset. 
$loss_{lat}$ and $loss_{mem}$ describe the losses caused by the hardware utilization which depend on the search-space configuration $\alpha$, $MCU_t$ and the user defined hardware limits $Lat_t$ and $Mem_t$.
The hardware loss functions are formulated as

\resizebox{\linewidth}{!}{
\begin{minipage}{\linewidth}
    \begin{align*}
        \text{loss}_{\text{lat}} &= \gamma_{\text{lat}} \cdot \log\left(\frac{\text{Lat}(\alpha, \text{MCU}_t)}{Lat_t}\right) \cdot [\text{Lat}(\alpha, \text{MCU}_t) \geq \text{Lat}_t] \\
        \text{loss}_{\text{mem}} &= \gamma_{\text{mem}} \cdot \log\left(\frac{\text{Mem}(\alpha, \text{MCU}_t)}{\text{Mem}_t}\right) \cdot [\text{Mem}(\alpha, \text{MCU}_t) \geq \text{Mem}_t]
    \end{align*}
    \label{eq:hw_loss}
\end{minipage}
}
The parameter $\gamma$ weights the importance of the individual loss-terms and needs to be set sufficiently high to ensure the user-defined hardware limits are not violated.
The complete search-algorithm can be seen in \autoref{alg:search} where both sets of weights are optimized in an iterative fashion.

\begin{algorithm}
        \caption{Search algorithm}\label{alg:search}
        \begin{algorithmic}[1]
            \FOR{$e \gets 1$ to $Epochs$} 
                \FOR{$b \gets 1$ to $Batches$} 
                \STATE \texttt{$\alpha = \alpha - \eta_1 \nabla_\alpha \mathcal{L}(\alpha, w, m, Lat_t, Mem_t)$}
                \STATE \texttt{$w = w - \eta_2 \nabla_w \mathcal{L}_{train}(\alpha, w)$}
                    \STATE $\tau$ $\gets$ $\tau \times \epsilon$ 
                \ENDFOR
            \ENDFOR
        \end{algorithmic}
\end{algorithm}

\section{Evaluation}
To showcase \systemName, we utilize two established benchmark datasets from the field of human activity recognition.
This section describes the evaluation of \systemName on two established benchmark datasets from the field of human activity recognition and displays the ability of \systemName to find suitable architectures under various latency and peak memory constraints.
The UCI-HAR dataset \cite{HAPT} features a window size of $128$ and nine sensor channels.
It was recorded with \SI{50}{\hertz} and features six classes.
The SkodaR dataset \cite{Skoda} features a window size of $64$ data points, $30$ sensor channels and was recorded with a sampling rate of \SI{96}{\hertz}.
Evaluation was performed on the \nucleof \cite{NUCLEOF473} equipped with an NRF52832 (ARM Cortex-M4, \SI{180}{\mega\hertz} CPU clock, \SI{512}{\kilo\byte} flash, \SI{128}{\kilo\byte} SRAM) and the NUCLEO-L552ZE-Q \cite{nucleo} equipped with a STM32L552 (ARM Cortex-M33, \SI{80}{\mega\hertz} CPU clock, \SI{512}{\kilo\byte} flash, \SI{256}{\kilo\byte} SRAM).

\subsection{Setup}
To demonstrate the ease of use of \systemName, the same hyperparameters were used for all the experiments.
For the convolutions in the Time-Reduce cells, $f_{max}$ was set to $16$ and $g$ was set to four.
For the Sensor-Fusion cells, $f_{max}$ was set to $64$ and $g$ to eight.
For the number of cells, $ts_{ml}$ was set to $16$, $ts_{ms}$ to five and $sf_s$ to two.
These settings were chosen to balance between the characterization cost to build the latency lookup table and search space flexibility.
For the search algorithm, we set $\epsilon$ to $0.995$, $\eta_{lat}$ to two and $\eta_{mem}$ to four.
With this setup the search-space for the UCI-HAR dataset contains $\approx 10^{13}$ architecture and $\approx 10^{22}$ for the SkodaR dataset.

\subsection{\systemName under different Computational Resource Constraints}

\subsubsection{Latency vs. Performance}
\begin{figure}
    \includegraphics[width=\linewidth]{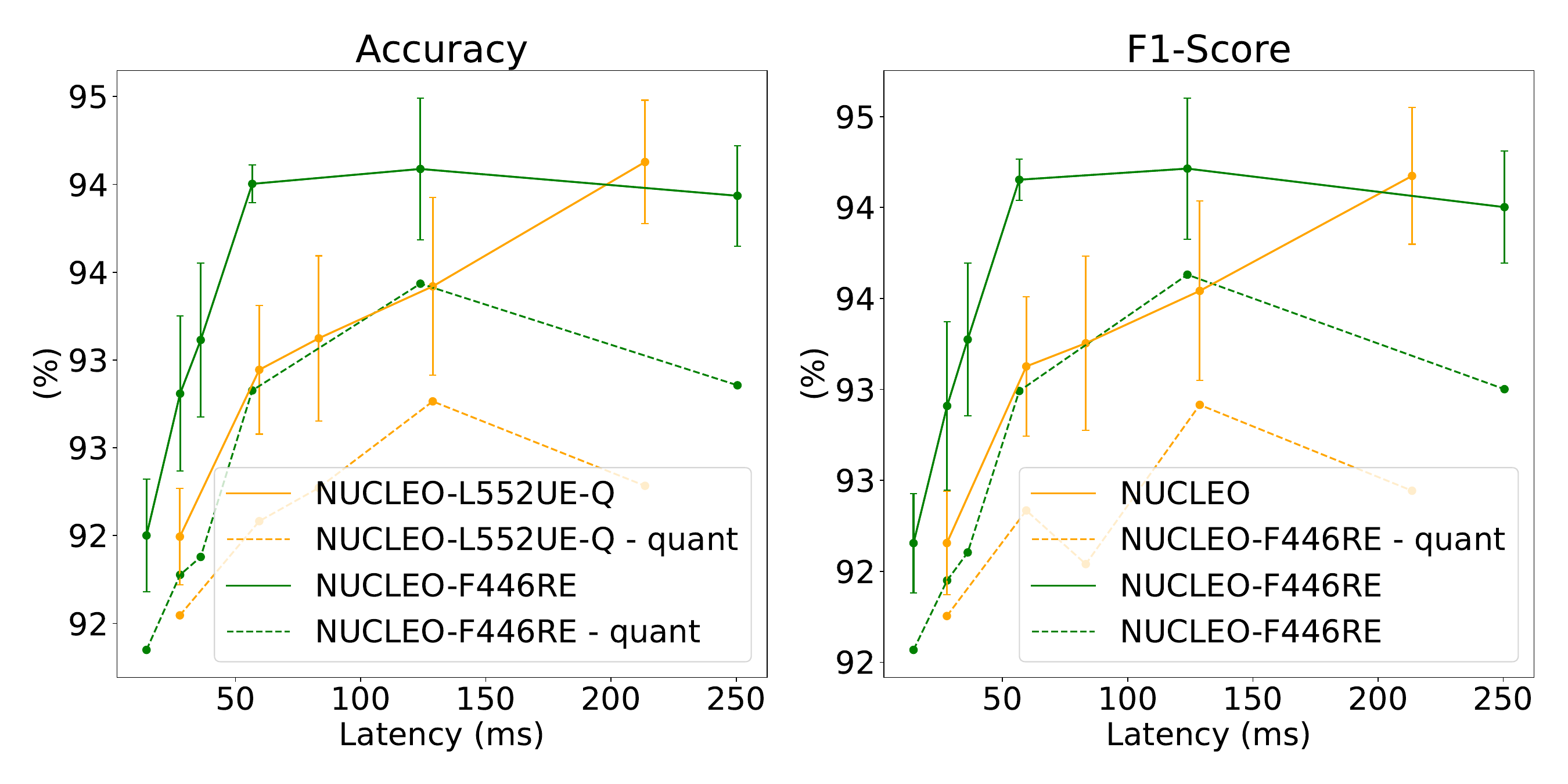}
    \caption{Trade-offs on the UCI-HAR dataset.
        Left: Accuracy, Right: F1-Score (Macro)}
    \label{fig:lat_ucihar}
\end{figure}

\begin{figure}
    \includegraphics[width=\linewidth]{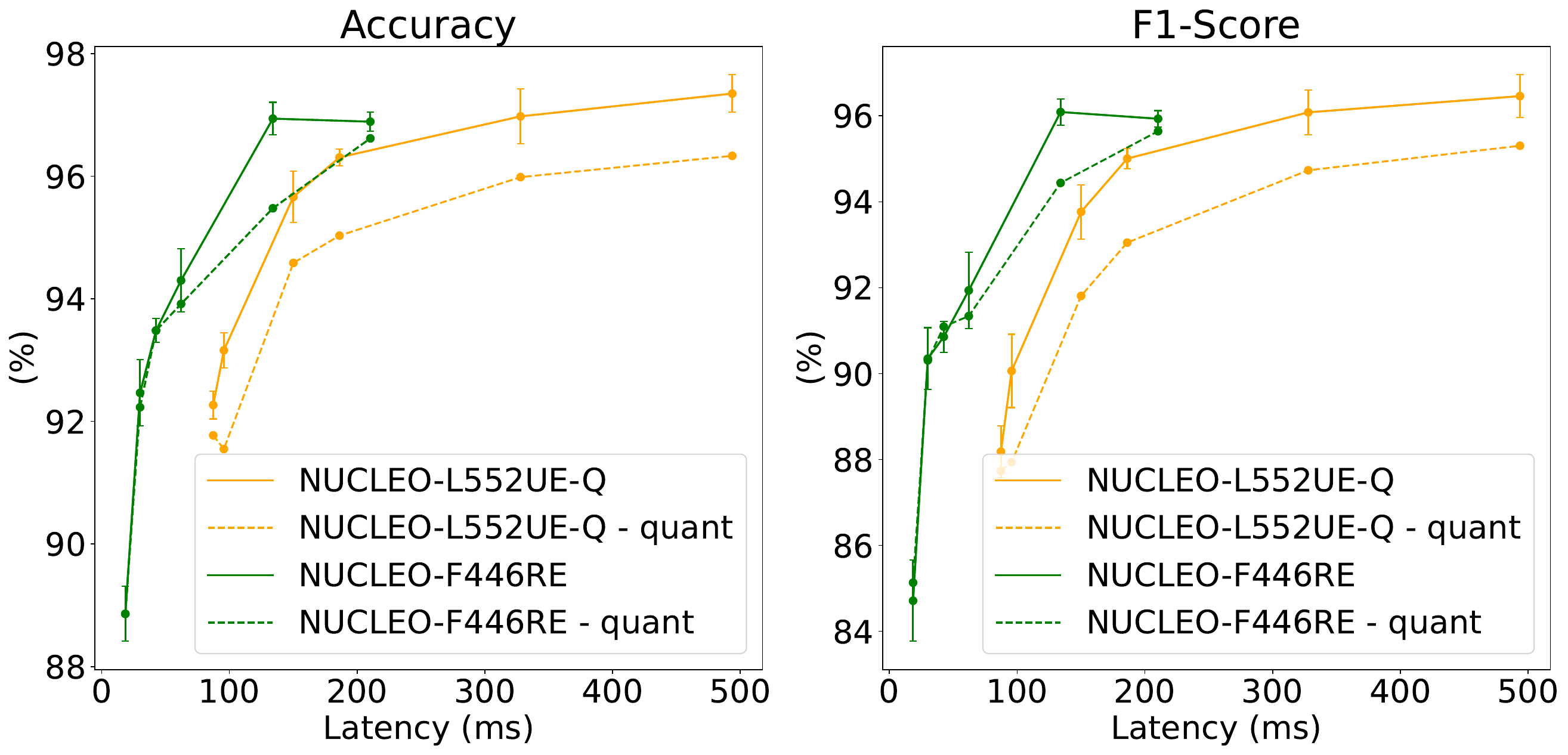}
    \caption{Trade-offs on the SkodaR dataset. Left: Accuracy, Right: F1-Score (Macro)}
    \label{fig:lat_skoda}
\end{figure}

This experiment demonstrates the ability of \systemName to find architectures under different latency constraints for which we disable the loss caused by the peak-memory consumption.
It is expected that the classification performance will increase as latency targets become higher which is also the cases, as seen in \autoref{fig:lat_ucihar} for the UCI-HAR dataset \cite{HAPT} and \autoref{fig:lat_skoda} for the SkodaR dataset \cite{Skoda}.
It can also be seen, that lower target latencies decrease performance more severely on the \nucleol as it is equipped with a weaker CPU.

\begin{table}
    \centering
    \caption{Three architectures found by \systemName for the SkodaR dataset compared to SOTA classifiers \cite{skoda_mahmud, DeepConvLSTM, tinyHar}.}
    \label{tab:comparison_skodaR}
    \begin{center}
        \begin{small}
            \begin{sc}
                \resizebox{\linewidth}{!}{
                \begin{tabular}{lccccccc}
                \toprule
                \textbf{Model} & \bf Device & \thead{\textbf{Latency} (ms) \\ \textbf{Peak memory} (B)} & \thead{ \bf Accuracy \\ non-quant \\ quant \\ (\%)} & \thead{\bf F1-Score \\ non-quant \\ quant \\(\%)} \\
                \midrule
                \systemName 1 & \nucleof  & \thead{30.07 \\ 21504} & \thead {92.47 \\ 92.23} & \thead {91.40 \\ 91.24} \\
                \systemName 2 & \nucleol  & \thead{150.09 \\ 19392} & \thead{95.66 \\ 94.58} & \thead {93.77 \\ 92.33} \\
                \systemName 3 & \nucleol  & \thead{493.69 \\ 34560} & \thead{97.35 \\ 96.33} & \thead {96.46 \\ 95.30} \\
                \midrule
                DeepConvLSTM & Desktop & \thead{1.1M \\ params} & - & \thead {98.99 \\ - } \\
                TinyHar & Desktop & \thead{67K \\ params} & - & \thead{98.82 \\ - } \\
                Mahmud et al. & Desktop & Not available* & - & \thead {97 \\ - } \\
                \midrule
                DARTS-softmax & Independent  & Not available* & failed & failed \\
                DARTS-gumbel & Independent & Not available* & \thead {96.84 \\ 95.04} & \thead
                {95.74 \\ 93.19} \\
                \bottomrule
                 \multicolumn{5}{l}{* Data not available from the source.}
                \end{tabular}}
                \end{sc}
        \end{small}
    \end{center}
\end{table}

\subsubsection{Peak Memory vs. Performance}
\begin{figure}
    \includegraphics[width=\linewidth]{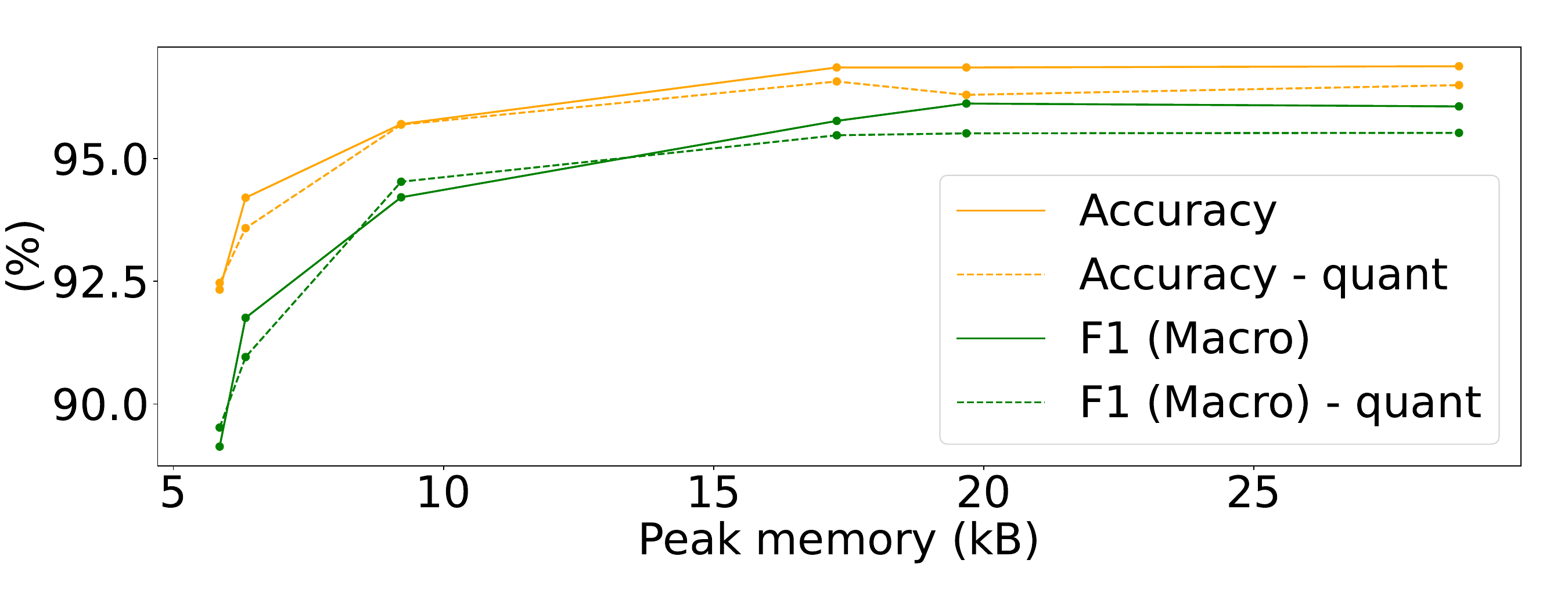}
    \caption{Trade-off between peak memory consumption and Accuracy / F1-Score (Macro). Comparison on the SkodaR dataset.}
    \label{fig:peak_mem_skodaR}
\end{figure}

This experiment demonstrates the ability of \systemName to find architectures under different peak memory constraints for which we disable the loss caused by the execution latency.
We expect performance to increase as more memory is allowed to be used which can be observed in \autoref{fig:peak_mem_skodaR}.
As the TFLM-framework \cite{tflm} is using the same amount of memory on every microcontroller, this experiment is independent of the microcontroller and is therefore only executed on the \nucleol \cite{nucleo}.

\begin{table}
    \centering
    \caption{Three architectures found by \systemName for the UCI-HAR dataset compared to SOTA classifiers \cite{uci_kolkar, uci_dua}. *: Data not available from the source.}
    \label{tab:comparison_ucihar}
        \begin{center}
        \begin{small}
            \begin{sc}
                \resizebox{\linewidth}{!}{
                \begin{tabular}{lcccccc}
                \toprule
                \textbf{Model} & \bf Device  & \thead{\textbf{Latency} (ms) \\ \textbf{Peak memory} (B)} & \thead{ \bf Accuracy \\ non-quant \\ quant \\ (\%)} & \thead{\bf F1-Score \\ non-quant \\ quant \\ (\%)} \\
                \midrule
                \systemName 1 & \nucleof & \thead{14.45 \\ 11520} & \thead{90.21 \\ 91.85} & \thead{92.66 \\ 92.07} \\
                \systemName 2 & \nucleof & \thead{123.84 \\ 15360} & \thead{94.59 \\ 93.93} & \thead{94.71 \\ 94.13} \\
                \systemName 3 & \nucleol & \thead{213.57 \\ 15360} & \thead{94.63 \\ 92.78} & \thead{94.67 \\ 92.94} \\
                \midrule
                Kolkar et al. & Desktop &  Not available* & \thead{96.83 \\ -} & - \\ 
                Dua et al. & Desktop & Not available* & \thead{96.20 \\ -} & \thead{96.19 \\ -} \\ 
                \midrule
                DARTS-softmax & Independent  & Not available* & \thead{94.08 \\ 92.01} & \thead{94.28 \\ 92.35} \\
                DARTS-gumbel & Independent & Not available* & \thead{95.40 \\ 92.97} & \thead{95.59 \\ 93.24} \\
                \bottomrule
                 \multicolumn{5}{l}{* Data not available from the source.}
                \end{tabular}}
            \end{sc}
        \end{small}
    \end{center}
\end{table}

\section{Discussion}
In comparison to existing systems, \systemName is the first to bring time-series classification to microcontrollers using neural architecture search in a hardware aware fashion.
As many IoT and wearable devices are equipped with a variety of time-series producing sensors, whose data must be processed, we expect many application scenarios to benefit from our presented methodology.
Especially when user data needs to be processed privately, in real-time or a connection to a server in the cloud is not feasible.

\subsection{Comparison to the State-of-the-Art}
To better understand the performance achieved by our system,  we evaluate \systemName on state-of-the-art benchmark datasets SkodaR \cite{Skoda} and UCI-HAR \cite{HAPT}.
\systemName is able to achieve performances closes to the state-of-the-art when comparing against time series classification systems found in the literature although these systems are running on desktop computers while neural networks found by \systemName are running on MCUs.
In addition, we compare \systemName to a DARTS-based baseline, where we replace the original search space with our own, to make it compatible to the time series classification task. For the $pseudo\_prob$-function we use the original Softmax implementation as well as the Gumbel-Softmax-function.
The results for the SkodaR dataset\cite{Skoda} can be found in \autoref{tab:comparison_skodaR} and for UCI-HAR \cite{HAPT} in \autoref{tab:comparison_ucihar}.
Further comparisons to other NAS-systems are not possible because they do not target the problem of time-series classification and therefore have incompatible search spaces.

\subsection{Limitations and Future Work}
Besides the neural network architecture, sampling rate, window size, and sensor selection can also impact classification performance \cite{samplingRate_study, window_size}. To create a complete end-to-end time series classification search system for MCUs, the proposed system can be expanded to encompass these parameters in the search space. %

\section{Conclusion}
This paper introduced \systemName, a first-of-its-kind hardware-aware neural architecture search (HW-NAS) system specifically designed for time series classification on resource constraint microcontrollers.
By utilizing two type of searchable cells, \systemName can be used for various datasets which differ in the window-length and the number of sensors.
This, coupled with the possibility to set limits on the execution latency and peak memory consumption makes the system usable in various application scenarios such as privacy critical or real-time systems.
The used lookup-table latency estimation approach allows to precisely calculate the execution latency of architectures in the search space and therefore enables \systemName to be used in real-time systems.
Our experimental results indicate, that for a variety of different hardware limits, \systemName is able to find a suitable neural network architecture while achieving classification performances close to state-of-the-art desktop models.

\newpage

\bibliographystyle{ACM-Reference-Format}
\bibliography{literature.bib}


\begin{thebibliography}{42}

%%% the Web address.

\ifx \showCODEN    \undefined \def \showCODEN     #1{\unskip}     \fi
\ifx \showDOI      \undefined \def \showDOI       #1{#1}\fi
\ifx \showISBNx    \undefined \def \showISBNx     #1{\unskip}     \fi
\ifx \showISBNxiii \undefined \def \showISBNxiii  #1{\unskip}     \fi
\ifx \showISSN     \undefined \def \showISSN      #1{\unskip}     \fi
\ifx \showLCCN     \undefined \def \showLCCN      #1{\unskip}     \fi
\ifx \shownote     \undefined \def \shownote      #1{#1}          \fi
\ifx \showarticletitle \undefined \def \showarticletitle #1{#1}   \fi
\ifx \showURL      \undefined \def \showURL       {\relax}        \fi
% invisible to TeX
\providecommand\bibfield[2]{#2}
\providecommand\bibinfo[2]{#2}
\providecommand\natexlab[1]{#1}
\providecommand\showeprint[2][]{arXiv:#2}

\bibitem[Abbas et~al\mbox{.}(2018)]%
        {healthMonitoring}
\bibfield{author}{\bibinfo{person}{Nasir Abbas}, \bibinfo{person}{Yan Zhang},
  \bibinfo{person}{Amir Taherkordi}, {and} \bibinfo{person}{Tor Skeie}.}
  \bibinfo{year}{2018}\natexlab{}.
\newblock \showarticletitle{Mobile Edge Computing: A Survey}.
\newblock \bibinfo{journal}{\emph{IEEE Internet of Things Journal}}
  \bibinfo{volume}{5}, \bibinfo{number}{1} (\bibinfo{year}{2018}),
  \bibinfo{pages}{450--465}.
\newblock
\urldef\tempurl%
\url{https://doi.org/10.1109/JIOT.2017.2750180}
\showDOI{\tempurl}


\bibitem[Abedin et~al\mbox{.}(2021)]%
        {attendAndDiscriminate}
\bibfield{author}{\bibinfo{person}{Alireza Abedin}, \bibinfo{person}{Mahsa
  Ehsanpour}, \bibinfo{person}{Qinfeng Shi}, \bibinfo{person}{Hamid
  Rezatofighi}, {and} \bibinfo{person}{Damith~C. Ranasinghe}.}
  \bibinfo{year}{2021}\natexlab{}.
\newblock \showarticletitle{Attend and Discriminate: Beyond the
  State-of-the-Art for Human Activity Recognition Using Wearable Sensors}.
\newblock \bibinfo{journal}{\emph{Proc. ACM Interact. Mob. Wearable Ubiquitous
  Technol.}} \bibinfo{volume}{5}, \bibinfo{number}{1}, Article
  \bibinfo{articleno}{1} (\bibinfo{date}{mar} \bibinfo{year}{2021}),
  \bibinfo{numpages}{22}~pages.
\newblock
\urldef\tempurl%
\url{https://doi.org/10.1145/3448083}
\showDOI{\tempurl}


\bibitem[Bagnall et~al\mbox{.}(2017)]%
        {tsc_bakeoff}
\bibfield{author}{\bibinfo{person}{Anthony Bagnall}, \bibinfo{person}{Jason
  Lines}, \bibinfo{person}{Aaron Bostrom}, \bibinfo{person}{James Large}, {and}
  \bibinfo{person}{Eamonn Keogh}.} \bibinfo{year}{2017}\natexlab{}.
\newblock \showarticletitle{The great time series classification bake off: a
  review and experimental evaluation of recent algorithmic advances}.
\newblock \bibinfo{journal}{\emph{Data Mining and Knowledge Discovery}}
  \bibinfo{volume}{31}, \bibinfo{number}{3} (\bibinfo{date}{01 May}
  \bibinfo{year}{2017}), \bibinfo{pages}{606--660}.
\newblock
\showISSN{1573-756X}
\urldef\tempurl%
\url{https://doi.org/10.1007/s10618-016-0483-9}
\showDOI{\tempurl}


\bibitem[Banos et~al\mbox{.}(2014)]%
        {window_size}
\bibfield{author}{\bibinfo{person}{Oresti Banos}, \bibinfo{person}{Juan-Manuel
  Galvez}, \bibinfo{person}{Miguel Damas}, \bibinfo{person}{Hector Pomares},
  {and} \bibinfo{person}{Ignacio Rojas}.} \bibinfo{year}{2014}\natexlab{}.
\newblock \showarticletitle{Window Size Impact in Human Activity Recognition}.
\newblock \bibinfo{journal}{\emph{Sensors}} \bibinfo{volume}{14},
  \bibinfo{number}{4} (\bibinfo{year}{2014}), \bibinfo{pages}{6474--6499}.
\newblock
\showISSN{1424-8220}
\urldef\tempurl%
\url{https://doi.org/10.3390/s140406474}
\showDOI{\tempurl}


\bibitem[Benmeziane et~al\mbox{.}(2021)]%
        {survey_benmeziane}
\bibfield{author}{\bibinfo{person}{Hadjer Benmeziane}, \bibinfo{person}{Kaoutar
  El~Maghraoui}, \bibinfo{person}{Hamza Ouarnoughi}, \bibinfo{person}{Smail
  Niar}, \bibinfo{person}{Martin Wistuba}, {and} \bibinfo{person}{Naigang
  Wang}.} \bibinfo{year}{2021}\natexlab{}.
\newblock \showarticletitle{Hardware-Aware Neural Architecture Search: Survey
  and Taxonomy}. In \bibinfo{booktitle}{\emph{Proceedings of the Thirtieth
  International Joint Conference on Artificial Intelligence, {IJCAI-21}}},
  \bibfield{editor}{\bibinfo{person}{Zhi-Hua Zhou}} (Ed.).
  \bibinfo{publisher}{International Joint Conferences on Artificial
  Intelligence Organization}, \bibinfo{pages}{4322--4329}.
\newblock
\urldef\tempurl%
\url{https://doi.org/10.24963/ijcai.2021/592}
\showDOI{\tempurl}
\newblock
\shownote{Survey Track}.


\bibitem[Brock et~al\mbox{.}(2017)]%
        {smash}
\bibfield{author}{\bibinfo{person}{Andrew Brock}, \bibinfo{person}{Theodore
  Lim}, \bibinfo{person}{James~M. Ritchie}, {and} \bibinfo{person}{Nick
  Weston}.} \bibinfo{year}{2017}\natexlab{}.
\newblock \showarticletitle{{SMASH:} One-Shot Model Architecture Search through
  HyperNetworks}.
\newblock \bibinfo{journal}{\emph{CoRR}}  \bibinfo{volume}{abs/1708.05344}
  (\bibinfo{year}{2017}).
\newblock
\showeprint[arXiv]{1708.05344}
\urldef\tempurl%
\url{http://arxiv.org/abs/1708.05344}
\showURL{%
\tempurl}


\bibitem[Cai et~al\mbox{.}(2019)]%
        {ProxylessNAS}
\bibfield{author}{\bibinfo{person}{Han Cai}, \bibinfo{person}{Ligeng Zhu},
  {and} \bibinfo{person}{Song Han}.} \bibinfo{year}{2019}\natexlab{}.
\newblock \bibinfo{title}{ProxylessNAS: Direct Neural Architecture Search on
  Target Task and Hardware}.
\newblock
\newblock
\showeprint[arxiv]{1812.00332}~[cs.LG]


\bibitem[Cao et~al\mbox{.}(2020)]%
        {predictiveMaintenance}
\bibfield{author}{\bibinfo{person}{Keyan Cao}, \bibinfo{person}{Yefan Liu},
  \bibinfo{person}{Gongjie Meng}, {and} \bibinfo{person}{Qimeng Sun}.}
  \bibinfo{year}{2020}\natexlab{}.
\newblock \showarticletitle{An Overview on Edge Computing Research}.
\newblock \bibinfo{journal}{\emph{IEEE Access}}  \bibinfo{volume}{8}
  (\bibinfo{year}{2020}), \bibinfo{pages}{85714--85728}.
\newblock
\urldef\tempurl%
\url{https://doi.org/10.1109/ACCESS.2020.2991734}
\showDOI{\tempurl}


\bibitem[Chen et~al\mbox{.}(2021)]%
        {tsNAS_chen}
\bibfield{author}{\bibinfo{person}{Donghui Chen}, \bibinfo{person}{Ling Chen},
  \bibinfo{person}{Zongjiang Shang}, \bibinfo{person}{Youdong Zhang},
  \bibinfo{person}{Bo Wen}, {and} \bibinfo{person}{Chenghu Yang}.}
  \bibinfo{year}{2021}\natexlab{}.
\newblock \bibinfo{title}{Scale-Aware Neural Architecture Search for
  Multivariate Time Series Forecasting}.
\newblock
\newblock
\urldef\tempurl%
\url{https://doi.org/10.48550/ARXIV.2112.07459}
\showDOI{\tempurl}


\bibitem[Chen and Ran(2019)]%
        {edgePrivacy}
\bibfield{author}{\bibinfo{person}{Jiasi Chen} {and} \bibinfo{person}{Xukan
  Ran}.} \bibinfo{year}{2019}\natexlab{}.
\newblock \showarticletitle{Deep Learning With Edge Computing: A Review}.
\newblock \bibinfo{journal}{\emph{Proc. IEEE}} \bibinfo{volume}{107},
  \bibinfo{number}{8} (\bibinfo{year}{2019}), \bibinfo{pages}{1655--1674}.
\newblock
\urldef\tempurl%
\url{https://doi.org/10.1109/JPROC.2019.2921977}
\showDOI{\tempurl}


\bibitem[Cui et~al\mbox{.}(2016)]%
        {multi_scale_cnn}
\bibfield{author}{\bibinfo{person}{Zhicheng Cui}, \bibinfo{person}{Wenlin
  Chen}, {and} \bibinfo{person}{Yixin Chen}.} \bibinfo{year}{2016}\natexlab{}.
\newblock \bibinfo{title}{Multi-Scale Convolutional Neural Networks for Time
  Series Classification}.
\newblock
\newblock
\urldef\tempurl%
\url{https://doi.org/10.48550/ARXIV.1603.06995}
\showDOI{\tempurl}


\bibitem[David et~al\mbox{.}(2021)]%
        {tflm}
\bibfield{author}{\bibinfo{person}{Robert David}, \bibinfo{person}{Jared Duke},
  \bibinfo{person}{Advait Jain}, \bibinfo{person}{Vijay Janapa~Reddi},
  \bibinfo{person}{Nat Jeffries}, \bibinfo{person}{Jian Li},
  \bibinfo{person}{Nick Kreeger}, \bibinfo{person}{Ian Nappier},
  \bibinfo{person}{Meghna Natraj}, \bibinfo{person}{Tiezhen Wang},
  \bibinfo{person}{Pete Warden}, {and} \bibinfo{person}{Rocky Rhodes}.}
  \bibinfo{year}{2021}\natexlab{}.
\newblock \showarticletitle{TensorFlow Lite Micro: Embedded Machine Learning
  for TinyML Systems}. In \bibinfo{booktitle}{\emph{Proceedings of Machine
  Learning and Systems}}, \bibfield{editor}{\bibinfo{person}{A.~Smola},
  \bibinfo{person}{A.~Dimakis}, {and} \bibinfo{person}{I.~Stoica}} (Eds.),
  Vol.~\bibinfo{volume}{3}. \bibinfo{pages}{800--811}.
\newblock
\urldef\tempurl%
\url{https://proceedings.mlsys.org/paper/2021/file/d2ddea18f00665ce8623e36bd4e3c7c5-Paper.pdf}
\showURL{%
\tempurl}


\bibitem[Dennis et~al\mbox{.}(2019)]%
        {ts_mcu_001}
\bibfield{author}{\bibinfo{person}{Don Dennis}, \bibinfo{person}{Durmus
  Alp~Emre Acar}, \bibinfo{person}{Vikram Mandikal},
  \bibinfo{person}{Vinu~Sankar Sadasivan}, \bibinfo{person}{Venkatesh
  Saligrama}, \bibinfo{person}{Harsha~Vardhan Simhadri}, {and}
  \bibinfo{person}{Prateek Jain}.} \bibinfo{year}{2019}\natexlab{}.
\newblock \showarticletitle{Shallow RNN: Accurate Time-series Classification on
  Resource Constrained Devices}. In \bibinfo{booktitle}{\emph{Advances in
  Neural Information Processing Systems}},
  \bibfield{editor}{\bibinfo{person}{H.~Wallach},
  \bibinfo{person}{H.~Larochelle}, \bibinfo{person}{A.~Beygelzimer},
  \bibinfo{person}{F.~d\textquotesingle Alch\'{e}-Buc},
  \bibinfo{person}{E.~Fox}, {and} \bibinfo{person}{R.~Garnett}} (Eds.),
  Vol.~\bibinfo{volume}{32}. \bibinfo{publisher}{Curran Associates, Inc.}
\newblock
\urldef\tempurl%
\url{https://proceedings.neurips.cc/paper_files/paper/2019/file/76d7c0780ceb8fbf964c102ebc16d75f-Paper.pdf}
\showURL{%
\tempurl}


\bibitem[Dua et~al\mbox{.}(2021)]%
        {uci_dua}
\bibfield{author}{\bibinfo{person}{Nidhi Dua}, \bibinfo{person}{Shiva~Nand
  Singh}, {and} \bibinfo{person}{Vijay~Bhaskar Semwal}.}
  \bibinfo{year}{2021}\natexlab{}.
\newblock \showarticletitle{Multi-input CNN-GRU based human activity
  recognition using wearable sensors}.
\newblock \bibinfo{journal}{\emph{Computing}} \bibinfo{volume}{103},
  \bibinfo{number}{7} (\bibinfo{date}{01 Jul} \bibinfo{year}{2021}),
  \bibinfo{pages}{1461--1478}.
\newblock
\showISSN{1436-5057}
\urldef\tempurl%
\url{https://doi.org/10.1007/s00607-021-00928-8}
\showDOI{\tempurl}


\bibitem[Ismail~Fawaz et~al\mbox{.}(2020)]%
        {InceptionTime}
\bibfield{author}{\bibinfo{person}{Hassan Ismail~Fawaz},
  \bibinfo{person}{Benjamin Lucas}, \bibinfo{person}{Germain Forestier},
  \bibinfo{person}{Charlotte Pelletier}, \bibinfo{person}{Daniel~F. Schmidt},
  \bibinfo{person}{Jonathan Weber}, \bibinfo{person}{Geoffrey~I. Webb},
  \bibinfo{person}{Lhassane Idoumghar}, \bibinfo{person}{Pierre-Alain Muller},
  {and} \bibinfo{person}{Fran{\c{c}}ois Petitjean}.}
  \bibinfo{year}{2020}\natexlab{}.
\newblock \showarticletitle{InceptionTime: Finding AlexNet for time series
  classification}.
\newblock \bibinfo{journal}{\emph{Data Mining and Knowledge Discovery}}
  \bibinfo{volume}{34}, \bibinfo{number}{6} (\bibinfo{date}{01 Nov}
  \bibinfo{year}{2020}), \bibinfo{pages}{1936--1962}.
\newblock
\showISSN{1573-756X}
\urldef\tempurl%
\url{https://doi.org/10.1007/s10618-020-00710-y}
\showDOI{\tempurl}


\bibitem[Jang et~al\mbox{.}(2017)]%
        {gumbel_softmax}
\bibfield{author}{\bibinfo{person}{Eric Jang}, \bibinfo{person}{Shixiang Gu},
  {and} \bibinfo{person}{Ben Poole}.} \bibinfo{year}{2017}\natexlab{}.
\newblock \showarticletitle{Categorical Reparameterization with
  Gumbel-Softmax}. In \bibinfo{booktitle}{\emph{5th International Conference on
  Learning Representations, {ICLR} 2017, Toulon, France, April 24-26, 2017,
  Conference Track Proceedings}}. \bibinfo{publisher}{OpenReview.net}.
\newblock
\urldef\tempurl%
\url{https://openreview.net/forum?id=rkE3y85ee}
\showURL{%
\tempurl}


\bibitem[Kim et~al\mbox{.}(2021)]%
        {samplingRate_study}
\bibfield{author}{\bibinfo{person}{Taehee Kim}, \bibinfo{person}{Jongman Kim},
  \bibinfo{person}{Bummo Koo}, \bibinfo{person}{Haneul Jung},
  \bibinfo{person}{Yejin Nam}, \bibinfo{person}{Yunhee Chang},
  \bibinfo{person}{Sehoon Park}, {and} \bibinfo{person}{Youngho Kim}.}
  \bibinfo{year}{2021}\natexlab{}.
\newblock \showarticletitle{Effects of Sampling Rate and Window Length on
  Motion Recognition Using sEMG Armband Module}.
\newblock \bibinfo{journal}{\emph{International Journal of Precision
  Engineering and Manufacturing}} \bibinfo{volume}{22}, \bibinfo{number}{8}
  (\bibinfo{year}{2021}), \bibinfo{pages}{1401--1411}.
\newblock
\showISBNx{2005-4602}
\urldef\tempurl%
\url{https://doi.org/10.1007/s12541-021-00546-6}
\showDOI{\tempurl}


\bibitem[Kolkar and Geetha(2021)]%
        {uci_kolkar}
\bibfield{author}{\bibinfo{person}{Ranjit Kolkar} {and} \bibinfo{person}{V.
  Geetha}.} \bibinfo{year}{2021}\natexlab{}.
\newblock \showarticletitle{Human Activity Recognition in Smart Home using Deep
  Learning Techniques}. In \bibinfo{booktitle}{\emph{2021 13th International
  Conference on Information \& Communication Technology and System (ICTS)}}.
  \bibinfo{pages}{230--234}.
\newblock
\urldef\tempurl%
\url{https://doi.org/10.1109/ICTS52701.2021.9609044}
\showDOI{\tempurl}


\bibitem[Lai et~al\mbox{.}(2018a)]%
        {CMSIS_NN}
\bibfield{author}{\bibinfo{person}{Liangzhen Lai}, \bibinfo{person}{Naveen
  Suda}, {and} \bibinfo{person}{Vikas Chandra}.}
  \bibinfo{year}{2018}\natexlab{a}.
\newblock \bibinfo{title}{CMSIS-NN: Efficient Neural Network Kernels for Arm
  Cortex-M CPUs}.
\newblock
\newblock
\urldef\tempurl%
\url{https://doi.org/10.48550/ARXIV.1801.06601}
\showDOI{\tempurl}


\bibitem[Lai et~al\mbox{.}(2018b)]%
        {not_all_ops_are_created_equal}
\bibfield{author}{\bibinfo{person}{Liangzhen Lai}, \bibinfo{person}{Naveen
  Suda}, {and} \bibinfo{person}{Vikas Chandra}.}
  \bibinfo{year}{2018}\natexlab{b}.
\newblock \bibinfo{title}{Not All Ops Are Created Equal!}
\newblock
\newblock
\urldef\tempurl%
\url{https://doi.org/10.48550/ARXIV.1801.04326}
\showDOI{\tempurl}


\bibitem[Liberis et~al\mbox{.}(2021)]%
        {uNAS_liberis}
\bibfield{author}{\bibinfo{person}{Edgar Liberis}, \bibinfo{person}{\L{}ukasz
  Dudziak}, {and} \bibinfo{person}{Nicholas~D. Lane}.}
  \bibinfo{year}{2021}\natexlab{}.
\newblock \showarticletitle{$\mu$NAS: Constrained Neural Architecture Search
  for Microcontrollers}. In \bibinfo{booktitle}{\emph{Proceedings of the 1st
  Workshop on Machine Learning and Systems}} (Online, United Kingdom)
  \emph{(\bibinfo{series}{EuroMLSys '21})}. \bibinfo{publisher}{Association for
  Computing Machinery}, \bibinfo{address}{New York, NY, USA},
  \bibinfo{pages}{70–79}.
\newblock
\showISBNx{9781450382984}
\urldef\tempurl%
\url{https://doi.org/10.1145/3437984.3458836}
\showDOI{\tempurl}


\bibitem[Liberis and Lane(2023)]%
        {liberies_pruning}
\bibfield{author}{\bibinfo{person}{Edgar Liberis} {and}
  \bibinfo{person}{Nicholas~D. Lane}.} \bibinfo{year}{2023}\natexlab{}.
\newblock \showarticletitle{Differentiable Neural Network Pruning to Enable
  Smart Applications on Microcontrollers}.
\newblock \bibinfo{journal}{\emph{Proc. ACM Interact. Mob. Wearable Ubiquitous
  Technol.}} \bibinfo{volume}{6}, \bibinfo{number}{4}, Article
  \bibinfo{articleno}{171} (\bibinfo{date}{jan} \bibinfo{year}{2023}),
  \bibinfo{numpages}{19}~pages.
\newblock
\urldef\tempurl%
\url{https://doi.org/10.1145/3569468}
\showDOI{\tempurl}


\bibitem[Liu et~al\mbox{.}(2019a)]%
        {TSCMultiConv}
\bibfield{author}{\bibinfo{person}{Chien-Liang Liu}, \bibinfo{person}{Wen-Hoar
  Hsaio}, {and} \bibinfo{person}{Yao-Chung Tu}.}
  \bibinfo{year}{2019}\natexlab{a}.
\newblock \showarticletitle{Time Series Classification With Multivariate
  Convolutional Neural Network}.
\newblock \bibinfo{journal}{\emph{IEEE Transactions on Industrial Electronics}}
  \bibinfo{volume}{66}, \bibinfo{number}{6} (\bibinfo{year}{2019}),
  \bibinfo{pages}{4788--4797}.
\newblock
\urldef\tempurl%
\url{https://doi.org/10.1109/TIE.2018.2864702}
\showDOI{\tempurl}


\bibitem[Liu et~al\mbox{.}(2019b)]%
        {darts}
\bibfield{author}{\bibinfo{person}{Hanxiao Liu}, \bibinfo{person}{Karen
  Simonyan}, {and} \bibinfo{person}{Yiming Yang}.}
  \bibinfo{year}{2019}\natexlab{b}.
\newblock \bibinfo{title}{DARTS: Differentiable Architecture Search}.
\newblock
\newblock
\showeprint[arxiv]{1806.09055}~[cs.LG]


\bibitem[Mahmud et~al\mbox{.}(2020)]%
        {skoda_mahmud}
\bibfield{author}{\bibinfo{person}{Saif Mahmud}, \bibinfo{person}{M.~T.~H.
  Tonmoy}, \bibinfo{person}{Kishor~Kumar Bhaumik}, \bibinfo{person}{A.~M.
  Rahman}, \bibinfo{person}{M.~A. Amin}, \bibinfo{person}{M. Shoyaib},
  \bibinfo{person}{Muhammad Asif~Hossain Khan}, {and} \bibinfo{person}{A.
  Ali}.} \bibinfo{year}{2020}\natexlab{}.
\newblock \showarticletitle{Human Activity Recognition from Wearable Sensor
  Data Using Self-Attention}. In \bibinfo{booktitle}{\emph{{ECAI} 2020 - 24th
  European Conference on Artificial Intelligence, 29 August-8 September 2020,
  Santiago de Compostela, Spain}}.
\newblock


\bibitem[Mendoza et~al\mbox{.}(2016)]%
        {towards_autoTune_networks_mendoza}
\bibfield{author}{\bibinfo{person}{Hector Mendoza}, \bibinfo{person}{Aaron
  Klein}, \bibinfo{person}{Matthias Feurer}, \bibinfo{person}{Jost~Tobias
  Springenberg}, {and} \bibinfo{person}{Frank Hutter}.}
  \bibinfo{year}{2016}\natexlab{}.
\newblock \showarticletitle{Towards Automatically-Tuned Neural Networks}. In
  \bibinfo{booktitle}{\emph{Proceedings of the Workshop on Automatic Machine
  Learning}} \emph{(\bibinfo{series}{Proceedings of Machine Learning Research},
  Vol.~\bibinfo{volume}{64})}, \bibfield{editor}{\bibinfo{person}{Frank
  Hutter}, \bibinfo{person}{Lars Kotthoff}, {and} \bibinfo{person}{Joaquin
  Vanschoren}} (Eds.). \bibinfo{publisher}{PMLR}, \bibinfo{address}{New York,
  New York, USA}, \bibinfo{pages}{58--65}.
\newblock
\urldef\tempurl%
\url{https://proceedings.mlr.press/v64/mendoza_towards_2016.html}
\showURL{%
\tempurl}


\bibitem[Microelectronics(2023)]%
        {NUCLEOF473}
\bibfield{author}{\bibinfo{person}{ST Microelectronics}.}
  \bibinfo{year}{2023}\natexlab{}.
\newblock \bibinfo{title}{NUCLEO-F446RE - STM32 Nucleo-64 development board
  with STM32F446RE MCU, supports Arduino and ST morpho connectivity -
  STMicroelectronics}.
\newblock
  \bibinfo{howpublished}{\url{https://www.st.com/en/evaluation-tools/nucleo-f446re.html}}.
\newblock
\newblock
\shownote{(Accessed on 08/14/2023)}.


\bibitem[Ordóñez and Roggen(2016)]%
        {DeepConvLSTM}
\bibfield{author}{\bibinfo{person}{Francisco~Javier Ordóñez} {and}
  \bibinfo{person}{Daniel Roggen}.} \bibinfo{year}{2016}\natexlab{}.
\newblock \showarticletitle{Deep Convolutional and LSTM Recurrent Neural
  Networks for Multimodal Wearable Activity Recognition}.
\newblock \bibinfo{journal}{\emph{Sensors}} \bibinfo{volume}{16},
  \bibinfo{number}{1} (\bibinfo{year}{2016}).
\newblock
\showISSN{1424-8220}
\urldef\tempurl%
\url{https://doi.org/10.3390/s16010115}
\showDOI{\tempurl}


\bibitem[Pham et~al\mbox{.}(2018)]%
        {enas}
\bibfield{author}{\bibinfo{person}{Hieu Pham}, \bibinfo{person}{Melody Guan},
  \bibinfo{person}{Barret Zoph}, \bibinfo{person}{Quoc Le}, {and}
  \bibinfo{person}{Jeff Dean}.} \bibinfo{year}{2018}\natexlab{}.
\newblock \showarticletitle{Efficient Neural Architecture Search via Parameters
  Sharing}. In \bibinfo{booktitle}{\emph{Proceedings of the 35th International
  Conference on Machine Learning}} \emph{(\bibinfo{series}{Proceedings of
  Machine Learning Research}, Vol.~\bibinfo{volume}{80})},
  \bibfield{editor}{\bibinfo{person}{Jennifer Dy} {and}
  \bibinfo{person}{Andreas Krause}} (Eds.). \bibinfo{publisher}{PMLR},
  \bibinfo{pages}{4095--4104}.
\newblock
\urldef\tempurl%
\url{https://proceedings.mlr.press/v80/pham18a.html}
\showURL{%
\tempurl}


\bibitem[Rakhshani et~al\mbox{.}(2020)]%
        {tsNas_rakhshani}
\bibfield{author}{\bibinfo{person}{Hojjat Rakhshani}, \bibinfo{person}{Hassan
  Ismail~Fawaz}, \bibinfo{person}{Lhassane Idoumghar}, \bibinfo{person}{Germain
  Forestier}, \bibinfo{person}{Julien Lepagnot}, \bibinfo{person}{Jonathan
  Weber}, \bibinfo{person}{Mathieu Brévilliers}, {and}
  \bibinfo{person}{Pierre-Alain Muller}.} \bibinfo{year}{2020}\natexlab{}.
\newblock \showarticletitle{Neural Architecture Search for Time Series
  Classification}. In \bibinfo{booktitle}{\emph{2020 International Joint
  Conference on Neural Networks (IJCNN)}}. \bibinfo{pages}{1--8}.
\newblock
\urldef\tempurl%
\url{https://doi.org/10.1109/IJCNN48605.2020.9206721}
\showDOI{\tempurl}


\bibitem[Rashid et~al\mbox{.}(2022)]%
        {humanActivityRecognition}
\bibfield{author}{\bibinfo{person}{Nafiul Rashid}, \bibinfo{person}{Berken~Utku
  Demirel}, {and} \bibinfo{person}{Mohammad Abdullah Al~Faruque}.}
  \bibinfo{year}{2022}\natexlab{}.
\newblock \showarticletitle{AHAR: Adaptive CNN for Energy-Efficient Human
  Activity Recognition in Low-Power Edge Devices}.
\newblock \bibinfo{journal}{\emph{IEEE Internet of Things Journal}}
  \bibinfo{volume}{9}, \bibinfo{number}{15} (\bibinfo{year}{2022}),
  \bibinfo{pages}{13041--13051}.
\newblock
\urldef\tempurl%
\url{https://doi.org/10.1109/JIOT.2022.3140465}
\showDOI{\tempurl}


\bibitem[Reyes-Ortiz et~al\mbox{.}(2016)]%
        {HAPT}
\bibfield{author}{\bibinfo{person}{Jorge-L. Reyes-Ortiz}, \bibinfo{person}{Luca
  Oneto}, \bibinfo{person}{Albert Samà}, \bibinfo{person}{Xavier Parra}, {and}
  \bibinfo{person}{Davide Anguita}.} \bibinfo{year}{2016}\natexlab{}.
\newblock \showarticletitle{Transition-Aware Human Activity Recognition Using
  Smartphones}.
\newblock \bibinfo{journal}{\emph{Neurocomputing}}  \bibinfo{volume}{171}
  (\bibinfo{year}{2016}), \bibinfo{pages}{754--767}.
\newblock
\showISSN{0925-2312}
\urldef\tempurl%
\url{https://doi.org/10.1016/j.neucom.2015.07.085}
\showDOI{\tempurl}


\bibitem[Sehrawat and Gill(2019)]%
        {smartSensors}
\bibfield{author}{\bibinfo{person}{Deepti Sehrawat} {and}
  \bibinfo{person}{Nasib~Singh Gill}.} \bibinfo{year}{2019}\natexlab{}.
\newblock \showarticletitle{Smart Sensors: Analysis of Different Types of IoT
  Sensors}. In \bibinfo{booktitle}{\emph{2019 3rd International Conference on
  Trends in Electronics and Informatics (ICOEI)}}. \bibinfo{pages}{523--528}.
\newblock
\urldef\tempurl%
\url{https://doi.org/10.1109/ICOEI.2019.8862778}
\showDOI{\tempurl}


\bibitem[STMicroelectronics(2023)]%
        {nucleo}
\bibfield{author}{\bibinfo{person}{STMicroelectronics}.}
  \bibinfo{year}{2023}\natexlab{}.
\newblock \bibinfo{title}{NUCLEO-L552ZE-Q - STM32 Nucleo-144 development
  board}.
\newblock
  \bibinfo{howpublished}{\url{https://www.st.com/en/evaluation-tools/nucleo-l552ze-q.html}}.
\newblock
\newblock
\shownote{(Accessed on 05/31/2022)}.


\bibitem[Wan et~al\mbox{.}(2020)]%
        {fbnetv2}
\bibfield{author}{\bibinfo{person}{Alvin Wan}, \bibinfo{person}{Xiaoliang Dai},
  \bibinfo{person}{Peizhao Zhang}, \bibinfo{person}{Zijian He},
  \bibinfo{person}{Yuandong Tian}, \bibinfo{person}{Saining Xie},
  \bibinfo{person}{Bichen Wu}, \bibinfo{person}{Matthew Yu},
  \bibinfo{person}{Tao Xu}, \bibinfo{person}{Kan Chen}, \bibinfo{person}{Peter
  Vajda}, {and} \bibinfo{person}{Joseph~E. Gonzalez}.}
  \bibinfo{year}{2020}\natexlab{}.
\newblock \showarticletitle{FBNetV2: Differentiable Neural Architecture Search
  for Spatial and Channel Dimensions}. In \bibinfo{booktitle}{\emph{2020
  IEEE/CVF Conference on Computer Vision and Pattern Recognition (CVPR)}}.
  \bibinfo{pages}{12962--12971}.
\newblock
\urldef\tempurl%
\url{https://doi.org/10.1109/CVPR42600.2020.01298}
\showDOI{\tempurl}


\bibitem[Wu et~al\mbox{.}(2019)]%
        {FBNet}
\bibfield{author}{\bibinfo{person}{Bichen Wu}, \bibinfo{person}{Xiaoliang Dai},
  \bibinfo{person}{Peizhao Zhang}, \bibinfo{person}{Yanghan Wang},
  \bibinfo{person}{Fei Sun}, \bibinfo{person}{Yiming Wu},
  \bibinfo{person}{Yuandong Tian}, \bibinfo{person}{Peter Vajda},
  \bibinfo{person}{Yangqing Jia}, {and} \bibinfo{person}{Kurt Keutzer}.}
  \bibinfo{year}{2019}\natexlab{}.
\newblock \showarticletitle{FBNet: Hardware-Aware Efficient ConvNet Design via
  Differentiable Neural Architecture Search}. In \bibinfo{booktitle}{\emph{2019
  IEEE/CVF Conference on Computer Vision and Pattern Recognition (CVPR)}}.
  \bibinfo{pages}{10726--10734}.
\newblock
\urldef\tempurl%
\url{https://doi.org/10.1109/CVPR.2019.01099}
\showDOI{\tempurl}


\bibitem[Yang and Zhang(2017)]%
        {ts_mcu_002}
\bibfield{author}{\bibinfo{person}{Fan Yang} {and} \bibinfo{person}{Lianyi
  Zhang}.} \bibinfo{year}{2017}\natexlab{}.
\newblock \showarticletitle{Real-time human activity classification by
  accelerometer embedded wearable devices}. In \bibinfo{booktitle}{\emph{2017
  4th International Conference on Systems and Informatics (ICSAI)}}.
  \bibinfo{pages}{469--473}.
\newblock
\urldef\tempurl%
\url{https://doi.org/10.1109/ICSAI.2017.8248338}
\showDOI{\tempurl}


\bibitem[Zappi et~al\mbox{.}(2012)]%
        {Skoda}
\bibfield{author}{\bibinfo{person}{Piero Zappi}, \bibinfo{person}{Daniel
  Roggen}, \bibinfo{person}{Elisabetta Farella}, \bibinfo{person}{Gerhard
  Troester}, {and} \bibinfo{person}{Luca Benini}.}
  \bibinfo{year}{2012}\natexlab{}.
\newblock \showarticletitle{Network-Level Power-Performance Trade-Off in
  Wearable Activity Recognition: A Dynamic Sensor Selection Approach}.
\newblock \bibinfo{journal}{\emph{ACM Transactions on Embedded Computing
  Systems}}  \bibinfo{volume}{11} (\bibinfo{date}{09} \bibinfo{year}{2012}),
  \bibinfo{pages}{68:1--68:30}.
\newblock
\urldef\tempurl%
\url{https://doi.org/10.1145/2345770.2345781}
\showDOI{\tempurl}


\bibitem[Zhang and Zhou(2021)]%
        {micronets}
\bibfield{author}{\bibinfo{person}{Shuai Zhang} {and} \bibinfo{person}{Xichuan
  Zhou}.} \bibinfo{year}{2021}\natexlab{}.
\newblock \showarticletitle{MicroNet: Realizing Micro Neural Network via
  Binarizing GhostNet}. In \bibinfo{booktitle}{\emph{2021 6th International
  Conference on Intelligent Computing and Signal Processing (ICSP)}}.
  \bibinfo{pages}{1340--1343}.
\newblock
\urldef\tempurl%
\url{https://doi.org/10.1109/ICSP51882.2021.9408972}
\showDOI{\tempurl}


\bibitem[Zhang et~al\mbox{.}(2023)]%
        {smart_sensors}
\bibfield{author}{\bibinfo{person}{Zixuan Zhang}, \bibinfo{person}{Luwei Wang},
  {and} \bibinfo{person}{Chengkuo Lee}.} \bibinfo{year}{2023}\natexlab{}.
\newblock \showarticletitle{Recent Advances in Artificial Intelligence
  Sensors}.
\newblock \bibinfo{journal}{\emph{Advanced Sensor Research}}
  \bibinfo{volume}{2}, \bibinfo{number}{8} (\bibinfo{year}{2023}),
  \bibinfo{pages}{2200072}.
\newblock
\urldef\tempurl%
\url{https://doi.org/10.1002/adsr.202200072}
\showDOI{\tempurl}
\showeprint{https://onlinelibrary.wiley.com/doi/pdf/10.1002/adsr.202200072}


\bibitem[Zhou et~al\mbox{.}(2022)]%
        {tinyHar}
\bibfield{author}{\bibinfo{person}{Yexu Zhou}, \bibinfo{person}{Haibin Zhao},
  \bibinfo{person}{Yiran Huang}, \bibinfo{person}{Till Riedel},
  \bibinfo{person}{Michael Hefenbrock}, {and} \bibinfo{person}{Michael Beigl}.}
  \bibinfo{year}{2022}\natexlab{}.
\newblock \showarticletitle{TinyHAR: A Lightweight Deep Learning Model Designed
  for Human Activity Recognition}. In \bibinfo{booktitle}{\emph{Proceedings of
  the 2022 ACM International Symposium on Wearable Computers}} (Cambridge,
  United Kingdom) \emph{(\bibinfo{series}{ISWC '22})}.
  \bibinfo{publisher}{Association for Computing Machinery},
  \bibinfo{address}{New York, NY, USA}, \bibinfo{pages}{89–93}.
\newblock
\showISBNx{9781450394246}
\urldef\tempurl%
\url{https://doi.org/10.1145/3544794.3558467}
\showDOI{\tempurl}


\bibitem[Zoph and Le(2017)]%
        {rl_nas_zoph}
\bibfield{author}{\bibinfo{person}{Barret Zoph} {and} \bibinfo{person}{Quoc~V.
  Le}.} \bibinfo{year}{2017}\natexlab{}.
\newblock \bibinfo{title}{Neural Architecture Search with Reinforcement
  Learning}.
\newblock
\newblock
\showeprint[arxiv]{1611.01578}~[cs.LG]


\end{thebibliography}

\newpage
\appendix
\onecolumn
\section{Peak Memory Consumption}
\label{sec:peak_mem}
For execution of the neural networks on the MCUs, we utilize the TensorFlow Light Micro Framework (TFLM) \cite{tflm} together with the CMSIS-NN kernel library \cite{CMSIS_NN} and $int\_8$ quantization.
To calculate the peak memory consumption of an architecture, the literature \cite{survey_benmeziane, micronets} proposes to use analytical estimation methods which is also the strategy used in this paper.
To execute an operation in a neural network, the input and output tensors of the operation need to be present in memory.
In addition to that, some operations, e.g. from the CMSIS-NN kernel library \cite{CMSIS_NN} require extra memory to perform the computation.
When calculating the peak memory of a sequential model without parallel connections, it can be computed as the maximum over the memory requirements of each operation.
In general, the required memory to perform an operation can be computed as stated in equation \ref{eq:memory_analytical_appendix}.

\begin{equation}
    op_{mem}(op) = \sum_{i} mem({input}_i) 
                        + mem(output) + extra\_{mem}(op). \label{eq:memory_analytical_appendix}
\end{equation}
The function $mem(x)$ calculates the memory required to store a tensor and considers the data format of it. When using $int\_8$ tensors, only one fourth of the memory is required in comparison to $float\_32$ tensors. The total memory required to run an operation can then be calculated by summing the memory requirements for the input and output tensors. On top of that some operations require extra memory to run which is considered in the calculation of the peak memory usage of an architecture \cite{CMSIS_NN}.
In addition, the TFLM-framework needs additional memory to execute a neural network, which is not taken into account by this system.

\section{Model Retraining and Quantization}
\label{sec:model_retraining_quantization}
During architecture search, weight sharing between architectures is applied which allows for an efficient search but at the same time prevents a single architecture to obtain its optimal weights.
Therefore, after an architecture has been found, this architecture is trained from scratch to achieve the maximum performance.
Training is performed in a quantization aware fashion as we later deploy the resulting model to an MCU using $int\_8$ quantization.
This greatly reduces computational cost on the microcontroller in terms of execution latency, peak memory consumption and storage requirements with only a minimal loss in classification performance.

\end{document}